\pdfoutput=1

\documentclass[11pt]{article}

\usepackage[]{emnlp2021}

\usepackage{times}
\usepackage{latexsym}
\usepackage{graphicx}
\usepackage{multirow}
\usepackage{amsmath,amsfonts,amssymb}

\usepackage[T1]{fontenc}

\usepackage[utf8]{inputenc}

\usepackage{microtype}

\title{TransferNet: An Effective and Transparent Framework for Multi-hop Question Answering over Relation Graph}

\author{Jiaxin Shi\textsuperscript{1,2} , Shulin Cao\textsuperscript{1} , Lei Hou\textsuperscript{1}\protect \thanks{\ \ Corresponding author.} , Juanzi Li\textsuperscript{1} and Hanwang Zhang\textsuperscript{3}  \\
  \textsuperscript{1}Department of Computer Science and Technology, BNRist, Tsinghua University, Beijing 100084, China \\
  \textsuperscript{2}Cloud BU, Huawei Technologies, \textsuperscript{3}Nanyang Technological University \\
  \tt { shijx12@gmail.com } \\
  \tt { \{caosl19@mails., houlei@, lijuanzi@\}tsinghua.edu.cn } \\
  \tt { hanwangzhang@ntu.edu.sg } \\
  }

\begin{document}
\maketitle
\begin{abstract}
Multi-hop Question Answering (QA) is a challenging task because it requires precise reasoning with entity relations at every step towards the answer. 
The relations can be represented in terms of labels in knowledge graph (\textit{e.g.}, \textit{spouse}) or text in text corpus (\textit{e.g.}, \textit{they have been married for 26 years}).
Existing models usually infer the answer by predicting the sequential relation path or aggregating the hidden graph features.
The former is hard to optimize, and the latter lacks interpretability.
In this paper, we propose TransferNet, an effective and transparent model for multi-hop QA, which supports both label and text relations in a unified framework.
TransferNet jumps across entities at multiple steps.
At each step, it attends to different parts of the question, computes activated scores for relations, and then transfer the previous entity scores along activated relations in a differentiable way.
We carry out extensive experiments on three datasets and demonstrate that TransferNet surpasses the state-of-the-art models by a large margin.
In particular, on MetaQA, it achieves 100\% accuracy in 2-hop and 3-hop questions.
By qualitative analysis, we show that TransferNet has transparent and interpretable intermediate results.
\end{abstract}

\section{Introduction}
Question answering (QA) plays a central role in artificial intelligence.
It requires machines to understand the free-form questions and infer the answers by analyzing information from a large corpus~\cite{rajpurkar2016squad,joshi2017triviaqa,chen2017reading} or structured knowledge base~\cite{bordes2015large,yih2015semantic,jiang2019freebaseqa}.
Along with the fast development of deep learning, especially the pretraining technology~\cite{devlin2018bert,lan2019albert}, state-of-the-art models have been shown comparative with human performance on simple questions that only need a single hop~\cite{petrochuk2018simplequestions,zhang2020retrospective}, \textit{e.g.}, \textit{Who is the CEO of Microsoft Corporation}.
However, multi-hop QA, which requires reasoning with the entity relations at multiple steps, is far from resolved~\cite{yang2018hotpotqa,dua2019drop,zhang2017variational,talmor2018web}.

\begin{figure}[t]
\includegraphics[width=\linewidth]{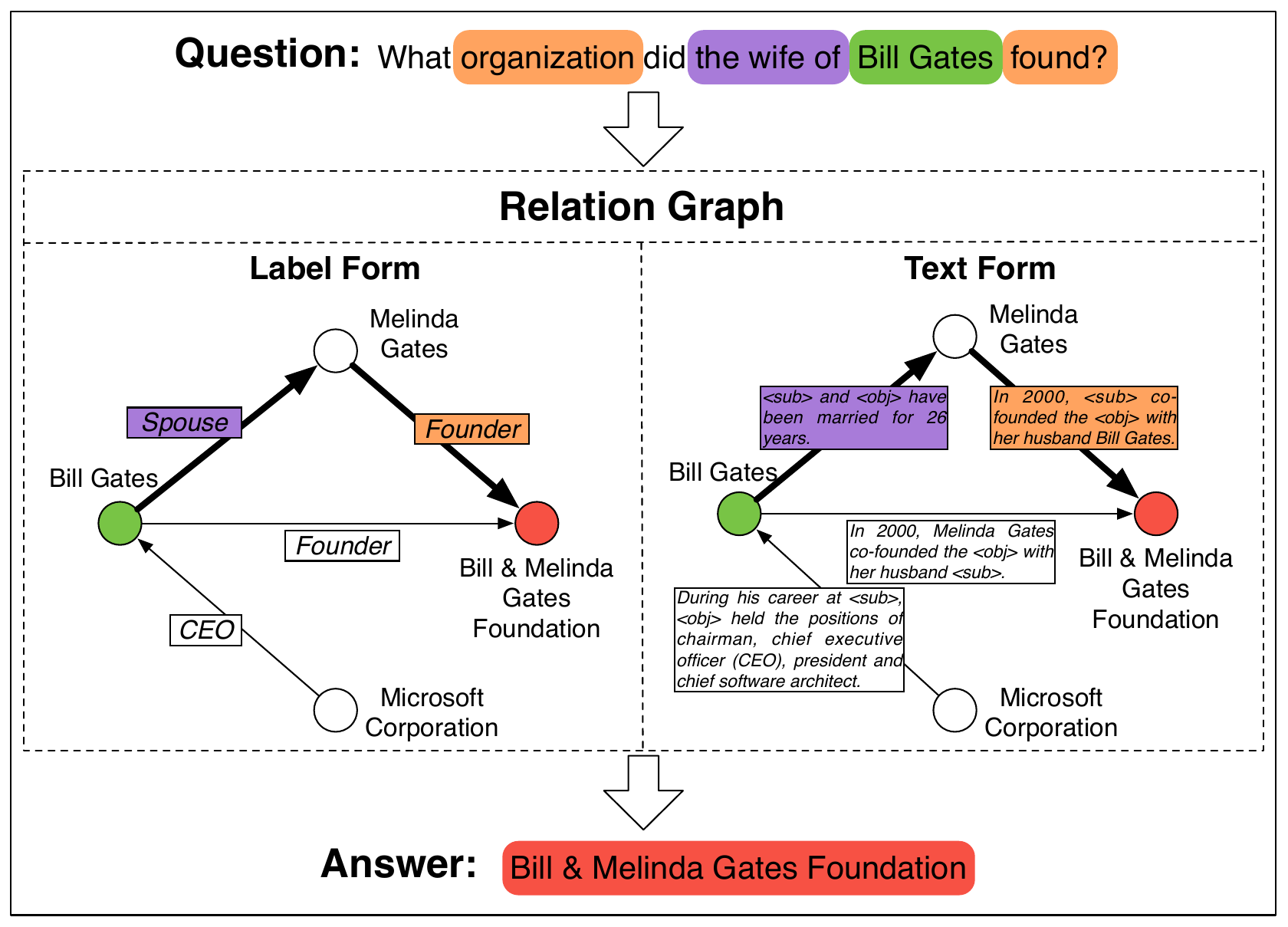}
\caption{Answering a multi-hop question over the relation graph. The relations are constrained predicates in the \textit{label form} (\textit{i.e.}, knowledge graph) while free texts in the \textit{text form}. The reasoning process has been marked in the graph, where the correspondence between relations and question words has been highlighted in the same color.}
\label{fig:intro}
\end{figure}

In this paper, we focus on multi-hop QA based on \textit{relation graphs}, which consists of entities and their relations.
As shown in Figure~\ref{fig:intro}, the relations can be represented by two forms: 
\begin{itemize}
    \item \textit{Label form}, also known as \textit{knowledge graph} (\textit{e.g.}, Freebase~\cite{Freebase}, Wikidata~\cite{Wikidata}), whose relations are manually-defined constrained predicates (\textit{e.g.}, \textit{Spouse}, \textit{CEO}).
    \item \textit{Text form}, whose relations are free texts retrieved from textual corpus. We can easily build the graph by extracting the co-occuring sentences of two entities. Since the label form is expensive and usually incomplete, the text form is more economical and practical.
\end{itemize}
In this paper, we aim to tackle multi-hop questions over these two different forms in a unified framework.

Existing methods for multi-hop QA have two main strands.
The first is to predict the sequential relation path in a weakly supervised setting~\cite{zhang2017variational,qiu2020stepwise}, that is, to learn the intermediate path only based on the final answer.
These works suffer from the convergence issues due to the huge search space, which heavily hinders their performance.
Besides, they are mostly proposed for the label form.
So, it is not clear how to adapt them to the text form, whose search space is even much huger.
The second strand is to collect evidences by using graph neural networks~\cite{sun2018open,sun2019pullnet}.
They can handle both the two relation forms and achieve state-of-the-art performance.
Although they prevail over the path-based models in performance, they are weak in interpretability since their intermediate reasoning process is black-box neural network layers.

In this paper, we propose a novel model for multi-hop QA, dubbed \textbf{TransferNet}, which has the following advantages:
1) \emph{Generality}. It can deal with the label form, the text form, and their combinations in a unified framework.
2) \emph{Effectiveness}. TransferNet outperforms previous models significantly, achieving 100\% accuracy of 2-hop and 3-hop questions in MetaQA dataset.
3) \emph{Transparency}. TransferNet is fully attention-based, so its intermediate steps can be easily visualized and understood by humans.

Specifically, TransferNet infers the answer by transfering entity scores along relation scores of multiple steps.
It starts from the topic entity of the question and maintains an entity score vector, whose elements indicate the probability of an entity being activated.
At each step, it attends to some question words (\textit{e.g.}, \textit{the wife of}) and compute scores for the relations in the graph.
Relations relevant to the question words will have high scores (\textit{e.g.}, \textit{Spouse}).
We formulate these relation scores into an adjacent matrix, where each entry indicates the transfer probability of an entity pair.
By multiplying the entity score vector with the relation score matrix, we can ``hop'' along relations in a differentiable manner.
After repeating for multiple steps, we can finally arrive at the target entity.

We conduct experiments for the two forms respectively.
For the label form, we use MetaQA~\cite{zhang2017variational}, WebQSP~\cite{yih2016value} and CompWebQ~\cite{talmor2018web}.
TransferNet achieves 100\% accuracy in the 2-hop and 3-hop questions of MetaQA.
On WebQSP and CompWebQ, we also achieve a significant improvement over state-of-the-art models.
For the text form, following \cite{sun2019pullnet}, we construct the relation graph of MetaQA from the WikiMovies corpus~\cite{miller2016key}.
We demonstrate that TransferNet surpasses previous models by a large margin, especially for the 2-hop and 3-hop questions.
When we mix the label form and the text form, TransferNet still keeps its superiority.
Moreover, by visualizing the intermediate results, we show its strong interpretability.
\footnote{\scriptsize\url{https://github.com/shijx12/TransferNet}}

\section{Related Work}
In this paper we focus on multi-hop question answering over the graph structure that is either knowledge graph or built from text corpus.
In previous works, GraftNet~\cite{sun2018open} and PullNet~\cite{sun2019pullnet} have a similar setting to ours but they mostly aim at the mixed form, which includes both label relations and text relations.
They first retrieve a question-specific subgraph and then use graph convolutional networks~\cite{kipf2016semi} to implicitly infer the answer entity.
These GCN-based methods are usually weak in interpretability because they cannot produce the intermediate reasoning path, which is necessary in our opinion for the task of multi-hop question answering.
Besides, there are many works specifically for only one graph form:

For the label form, which is also known as ``KBQA'' or ``KGQA'', existing methods fall into two categories: information retrieval~\cite{miller2016key,xu2019enhancing,zhao2019simple,saxena2020improving} and semantic parsing~\cite{berant2013semantic,yih2015semantic,liang2017neural,guo2018dialog,saha2019complex}.
The former retrieves answer from KG by learning representations of question and graph, while the latter queries answer by parsing the question into logical form.
Among these methods, VRN~\cite{zhang2017variational} and SRN~\cite{qiu2020stepwise} have a good interpretability as they learn an explicit reasoning path with reinforcement learning.
However, they suffer from the convergency issue due to the huge search space.
IRN~\cite{zhou2018interpretable} and ReifKB~\cite{cohen2020scalable} learn a soft distribution for intermediate relations and can be optimized using only the final answer.
However, it is not clear how to extend them to the text form.

Question answering over text corpus is also known as ``reading comprehension''.
For simple questions, whose answer can be retrieved directly from the text, pretrained models~\cite{devlin2018bert,lan2019albert} have performed better than humans~\cite{zhang2020retrospective}.
For multi-hop questions that are much more challenging, existing works~\cite{ding2019cognitive,fang2019hierarchical,tu2020select,zhao2019transformer} usually convert the text into a rule-based or learning-based entity graph, and then use graph neural networks~\cite{kipf2016semi} to perform implicit reasoning.
Similar to PullNet, they are weak in interpretability.
Besides, most of them build the graph by just connecting relevant entities, missing the important edge textual information.

\section{Methodology}

\begin{figure*}[t]
\includegraphics[width=\linewidth]{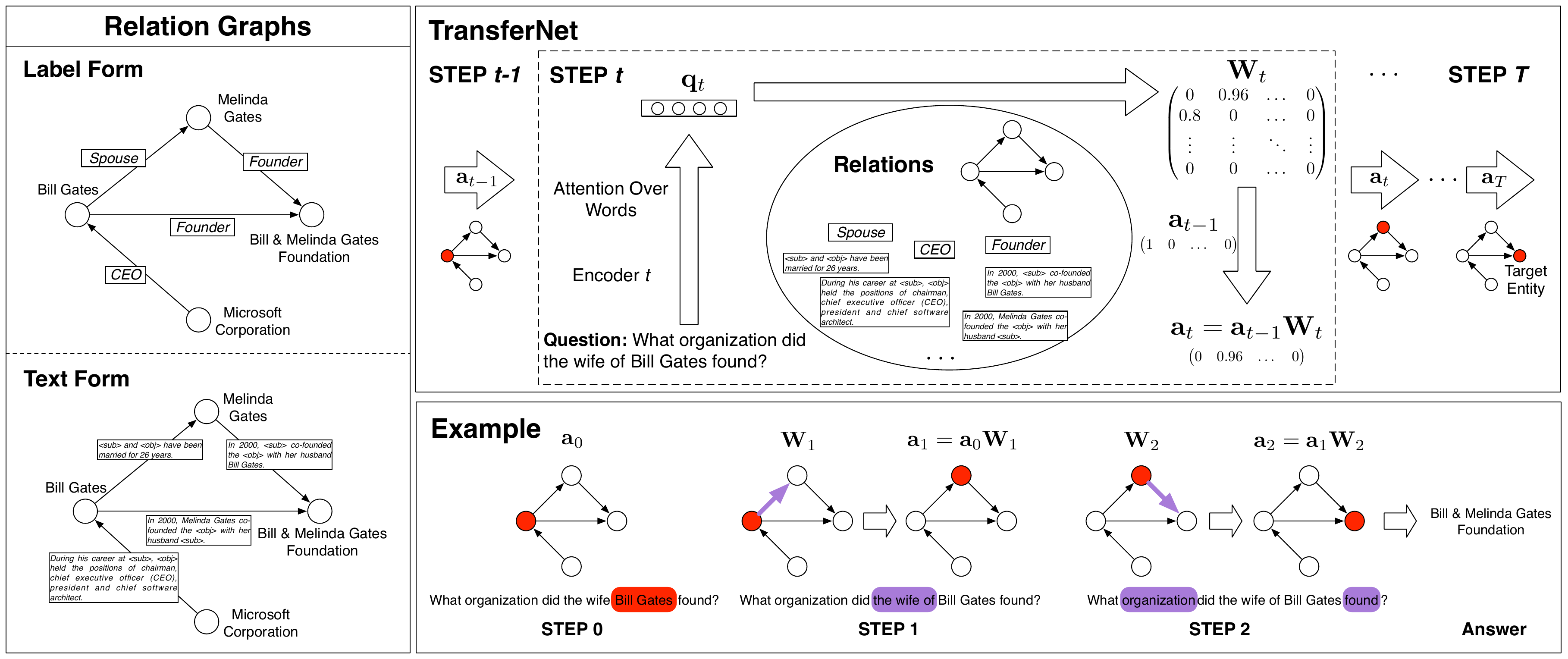}
\caption{The framework of TransferNet (top) and example of reasoning process (bottom).}
\label{fig:model}
\end{figure*}

\subsection{Preliminary}
We conduct multi-hop reasoning on a \textbf{relation graph}, which takes entities as nodes and relations between them as edges.
The relations can be of different forms, specifically, \emph{constrained labels} or \emph{free texts}.
The former is also known as structured Knowledge Graph (\textit{e.g.}, Wikidata~\cite{Wikidata}), which predefines a set of predicates to represent the entity relations.
The latter can be easily extracted from large-scale document corpora according to the co-occurence of entity pairs.
Figure~\ref{fig:intro} shows examples of these two forms.
In this paper we call them \textbf{label form} and \textbf{text form} respectively, and use \textbf{mixed form} to denote a relation graph consisting of both labels and texts.

We denote a relation graph as $\mathcal{G}$, its entities as $\mathcal{E}$ and its edges as $\mathcal{R}$.
Let $n$ denote the number of entities, then $\mathcal{R}$ is an $n \times n$ matrix whose element $r_{i,j}$ represents the relations between the head entity $e_i$ and the tail entity $e_j$.
$r_{i,j}$ can be a set of labels (for label form) or texts (for text form) or both (for mixed form).
A multi-hop question $q$ usually starts from a topic entity $e_x$ and needs to traverse across relations to reach the answer entities $Y=\{e_{y^1}, \cdots, e_{y^{|Y|}}\}$.

\subsection{TransferNet}
To infer the answer of a multi-hop question, TransferNet starts from the topic entity and jumps for $T$ steps.
At each step, it attends to different parts of the question to determine the most proper relation.
TransferNet maintains a score for each entity to denote their activated probabilities, which are initialized to 1 for the topic entity and 0 for the others.
At each step, TransferNet computes a score for each relation to denote their activated probabilities in terms of the current query, and then transfer the entity scores across those activated relations.
Figure~\ref{fig:model} shows the framework.

Formally, we denote the entity scores of step $t$ as a row vector $\mathbf{a}^t \in [0,1]^n$, where $[0,1]$ means a real number between $0$ and $1$.
$\mathbf{a}^0$ is the initial scores, \textit{i.e.}, only the topic entity $e_x$ gets $1$.
At step $t$, we attend to part of the question to get the query vector $\mathbf{q}^t \in \mathcal{R}^d$, where $d$ is the hidden dimension.
\begin{equation}
\begin{aligned}
&\mathbf{q}, (\mathbf{h}_1, \cdots, \mathbf{h}_{|q|}) = \text{Encoder}(q; \theta_e), \\
&\mathbf{qk}^t = f^t (\mathbf{q}; \theta_{f^t}), \\
&\mathbf{b}^t = \text{Softmax} (\mathbf{qk}^t \cdot [\mathbf{h}_1; \cdots; \mathbf{h}_{|q|}]^\top), \\
&\mathbf{q}^t = \sum_{i=1}^{|q|} b^t_i \mathbf{h}_i.
\end{aligned}
\end{equation}
$\mathbf{q}$ denotes the question embedding. $f^t$ is a projecting function of step $t$, which maps $\mathbf{q}$ to a specific query key $\mathbf{qk}^t$. $\mathbf{qk}^t$ is the attention key to compute scores for each word based on their hidden vector $\mathbf{h}_i$. $\mathbf{q}^t$ is the weighted sum of $\mathbf{h}_i$.

In terms of $\mathbf{q}^t$ TransferNet computes the relation scores $\mathbf{W}^t \in [0,1]^{n \times n}$:
\begin{equation}\label{eq:relation_score}
\mathbf{W}^t = g(\mathbf{q}^t; \theta_g).
\end{equation}
$\theta_g$ denotes the learnable parameters. 
We will have different implementations of $g$ for the label form and the text form, which will be introduced in Sec.\ref{sec:two_forms}.

Then we can simulate the ``jumping across edges'' as the following formulation:
\begin{equation}
\mathbf{a}^{t} = \mathbf{a}^{t-1} \mathbf{W}^{t}.
\end{equation}
Specifically, we have
\begin{equation}\label{eq:a_multi_W}
a^t_j = \sum_{i=1}^n a^{t-1}_i \times W^t_{i,j}.
\end{equation}
It means that the production of entity $e_i$'s previous score and the edge $r_{i,j}$'s current score will be collected into $e_j$'s current score.

After repeating for $T$ times, we get the entity scores of each step $\mathbf{a}^1, \mathbf{a}^2, \cdots, \mathbf{a}^T$.
Then we compute their weighted sum as the final output:
\begin{equation}
\begin{aligned}
\mathbf{c} &= \text{Softmax}(\text{MLP}(\mathbf{q})),\\
\mathbf{a}^* &= \sum_{t=1}^{T} c_t \mathbf{a}^t,
\end{aligned}
\end{equation}
where $\mathbf{c} \in [0,1]^T$ denotes the probability distribution of the question's hop, and $c_t$ is the probability value of hop $t$.
We can answer all questions from $1$-hop to $T$-hop by automatically determine its hop number.
The entity with maximum score in $\mathbf{a}^*$ is outputed as the answer.

TransferNet is a highly-transparent model.
As shown in the example of Figure~\ref{fig:model}, we can easily track the model behaviour by visualizing the activated words, relations, and entities at each step (see Sec.\ref{sec:interpretability} for more examples).

\subsection{Training}
Given the golden answer set $Y=\{e_{y^1}, \cdots, e_{y^{|Y|}}\}$, we construct the target score vector $\mathbf{y} \in \{0,1\}^n$ by
\begin{equation}
y_i = \left\{
    \begin{aligned}
    1, \ &\text{if  } e_i \in Y,\\
    0, \ &\text{else}.
    \end{aligned}
\right.
\end{equation}
Then we take the L2 Euclidean distance between $\mathbf{a}^*$ and $\mathbf{y}$ as our training objective:
\begin{equation}\label{eq:objective}
\mathcal{L} = \lVert \mathbf{a}^* - \mathbf{y} \rVert.
\end{equation}

Note that TransferNet is totally differentiable, therefore we can learn all of the intermediate scores (\textit{i.e.}, question attention, relation scores, and entity scores of each step) via this simple objective..

\subsection{Additional Modules}
We propose two modules to facilitate the learning of TransferNet.

\noindent\textbf{Score Truncation.}
According to Equation~\ref{eq:a_multi_W}, $a^t_j$ may exceed $1$ after a transfer step.
A too large score will have a bad influence to the gradient computation.
Especially when the hop increases, it may lead to gradient explosion.
Besides, our loss function, Equation~\ref{eq:objective}, will fail if the final score has an unlimited value.
So we need to rectify the entity scores after each transfer step, to ensure the value range is in $[0,1]$.
At the same time, we need to maintain the differentiability of the operation.
We propose such a truncation function:
\begin{equation}
\begin{aligned}
\text{Trunc}(a) &= a/z(a),\\
z(a) &=  \left\{
    \begin{aligned}
    &a.\text{detach()}, &\text{if  } a > 1, \\
    &1, &\text{if  } a \le 1.
    \end{aligned}
\right.
\end{aligned}
\end{equation}
After each transfer step, we truncate $\mathbf{a}^{t}$ by applying this function to each of its elements.

\noindent\textbf{Language Mask.}\label{sec:language_mask}
TranferNet does not consider the language bias of the question, which may include some hints for its answer.
For example, in the text-formed relation graph we may have (\textit{Harry Potter}, \textit{<sub> was published in <obj>}, \textit{United Kingdom}) and (\textit{Harry Potter}, \textit{<sub> was published in <obj>}, \textit{1997}). 
These two triples depict different aspects (\textit{i.e.}, the publication place and the publication time of \textit{Harry Potter}) but with the same relation text.
As a result, given the question \textit{Where was Harry Potter published}, TransferNet will produce the same scores for \textit{United Kingdom} and \textit{1997}, and thus use \textit{1997} to wrongly answer the \textit{Where}-question. 

To solve this issue, we propose a language mask to incorporate the question hints.
We predict a mask score for each entity using the question embedding:
\begin{equation}
\mathbf{m} = \text{Sigmoid}(\text{MLP}(\mathbf{q})),
\end{equation}
where $\mathbf{m} \in [0,1]^n$, $m_i$ denotes the mask score of entity $e_i$, MLP (short for multi-layer perceptron) projects $d$-dimensional feature to $n$-dimension.
We multiply the mask to the final entity scores,
\begin{equation}
\hat{\mathbf{a}^{*}} = \mathbf{m} \odot \mathbf{a}^*,
\end{equation}
where $\odot$ means element-wise multiplication.
The $\mathbf{a}^*$ in the objective function Equation~\ref{eq:objective} should be replaced with $\hat{\mathbf{a}^{*}}$.
Note that we need the language mask only in the text form, because the predicates of label form have no ambiguity.

\subsection{Relation Score Computation}\label{sec:two_forms}
Consider Equation~\ref{eq:relation_score}, $\mathbf{W}^t = g(\mathbf{q}^t; \theta_g)$, we design different implementations of $g$ for different relation forms.

\subsubsection{Label Form}
In the label form, relations are represented with a fixed predicate set $\mathcal{P}$.
We first compute probabilities for these predicates in terms of $\mathbf{q}^t$, and then collect corresponding probabilities of $r_{i,j}$ as $W^t_{i,j}$.

Formally, the predicate distribution is computed by
\begin{equation}
\mathbf{p}^t = \text{Softmax}(\text{MLP}(\mathbf{q}^t)).
\end{equation}
The Softmax function can be replaced with Sigmoid if predicates are not mutually exclusive, \textit{i.e.}, multiple predicates will be activated meanwhile.
Let $b$ denote the maximum number of relations between a pair of entity, then we can denote the relation as $r_{i,j} = \{r_{i,j,1}, \cdots, r_{i,j,b}\}$, where $r_{i,j,k} \in \{1, 2, \cdots, |\mathcal{P}|\}$.
The predicate probabilities are collected in terms of the relation labels:
\begin{equation}
W^t_{i,j} = \sum_{k=1}^{b} p^t_{r_{i,j,k}}.
\end{equation}
We gather the probabilities by summing them up.
$\max$ is another feasible option, but we find $\sum$ is more efficient and more stable.

\subsubsection{Text Form}\label{sec:text_form}
In the text form, relations are represented with natural language descriptions.
The graph is built by extracting the co-occuring sentence of a pair of entity and replacing the entities with special placeholders.
For example, the sentence \textit{Bill Gates and Melinda Gates have been married for 26 years} contributes an edge from \textit{Bill Gates} to \textit{Melinda Gates}, whose relation text is \textit{<sub> and <obj> have been married for 26 years}, as shown in Figure~\ref{fig:model}.
We can get the reverse relations by exchanging the placeholders of subject and object, but for simplicity, we do not show them in the figure.

Let $r_{i,j}=\{r_{i,j,1}, \cdots, r_{i,j,b}\}$ and $r_{i,j,k}$ denotes the $k$-th relation sentence.
We use a relation encoder to obtain the relation embeddings, and then compute the relation score by
\begin{equation}
\begin{aligned}
\mathbf{r}_{i,j,k} &= \text{Encoder}(r_{i,j,k}; \theta_r),\\
p^t_{r_{i,j,k}} &= \text{Sigmoid}(\text{MLP}(\mathbf{r}_{i,j,k} \odot \mathbf{q}^t)),\\
W^t_{i,j} &= \sum_{k=1}^{b} p^t_{r_{i,j,k}},
\end{aligned}
\end{equation}
where $\odot$ means element-wise product, MLP maps the feature from $d$-dimensional to $1$-dimensional.

Since there are a huge amount of (usually millions of) relation texts in a relation graph, it is impossible to compute the embeddings and scores for all of them.
So in practice, we select a subset of relations at each step.
Specifically, at step $t$, we select entities whose previous score $a^{t-1}_i$ is larger than a predefined threshold $\tau$ and only consider relations that start from these entities.
Besides, if there are too many relations meeting this condition, we will only preserve top $\omega$ of them, sorting based on their subject entity score.
By doing so, we just need to consider at most $\omega$ relations at each step.

We use the same method to process the mixed form, by simply regarding the label predicates as one-word sentences.

\section{Experiments}

\subsection{Datasets}
\noindent\textbf{MetaQA}~\cite{zhang2017variational} is a large-scale dataset of multi-hop question answering over knowledge graph, which extends WikiMovies~\cite{miller2016key} from single-hop to multi-hop.
It contains more than 400k questions, which are generated using dozens of templates and have up to 3 hops.
Its knowledge graph is from the movie domain, including 43k entities, 9 predicates, and 135k triples.

Besides the label from, we also constructed the text form of MetaQA by extracting the text corpus of WikiMovies~\cite{miller2016key}, which introduces the information of movies with free text.
Following \cite{sun2019pullnet}, we used exact match of surface forms for entity recognition and linking.
Given an article of a movie, we took the movie as subject and the other relavant entities (\textit{e.g.}, mentioned actor, year, and etc) as objects.
The sentence was processed with placeholders, that is, replacing the movie with \textit{<sub>} (if it occurs) and the object entity with \textit{<obj>}, and then regarded as the relation texts.
An entity pair can have multiple textual relations.

\noindent\textbf{WebQSP}~\cite{yih2016value} has a smaller scale of questions but larger scale of knowledge graph.
It contains thousands of natural language questions based on Freebase~\cite{Freebase}, which has millions of entities and triples.
Its questions are either 1-hop or 2-hop.
Following \cite{saxena2020improving}, we pruned the knowledge base to contain only mentioned predicates and within 2-hop triples of mentioned entities.
As a result, the processed knowledge graph includes 1.8 million entities, 572 predicates, and 5.7 million triples.
We only consider the label form of WebQSP due to its huge scale.

\noindent\textbf{CompWebQ}~\cite{talmor2018web} is an extended version of WebQSP with more hops and constraints.
Following \cite{sun2019pullnet}, we retrieved a subgraph for each question using PageRank algorithm.
On average, there are 1948 entities in each subgraph and the recall is 64\%.
Table~\ref{tab:data_statistics} lists the statistics of these datasets.

\begin{table}[ht]
\small
\centering
\begin{tabular}{l|ccc}
\hline
\textbf{Dataset} & \textbf{Train} & \textbf{Dev} & \textbf{Test} \\
\hline
MetaQA 1-hop & 96,106 & 9,992 & 9,947 \\
MetaQA 2-hop & 118,948 & 14,872 & 14,872 \\
MetaQA 3-hop & 114,196 & 14,274 & 14,274 \\
WebQSP & 2,998 & 100 & 1,639 \\
CompWebQ & 27,623  &  3,518  &  3,531 \\
\hline
\end{tabular}
\caption{\label{tab:data_statistics} Dataset statistics. }
\end{table}

\subsection{Baselines}
\noindent\textbf{KVMemNN}~\cite{miller2016key} uses the key-value memory to store knowledge and conducts multi-hop reasoning by iteratively reading the memory.

\noindent \textbf{VRN}~\cite{zhang2017variational} learns the reasoning path via reinforcement learning. Its intermediate results have a good interpretability.

\noindent \textbf{SRN}~\cite{qiu2020stepwise} improves VRN by beam search and reward shaping strategy, boosting its speed and performance.

\noindent \textbf{GraftNet}~\cite{sun2018open} extracts a question-specific subgraph from the entire relation graph with heuristics, and then uses graph neural networks to infer the answer.

\noindent \textbf{PullNet}~\cite{sun2019pullnet} improves GraftNet by learning to retrieve the subgraph with a graph CNN instead of heuristics.

\noindent \textbf{ReifKB}~\cite{cohen2020scalable} proposes a scalable implementation of probability transfer over large-scale knowledge graph of label form. It can be regarded as a degenerated case of TransferNet.

\noindent \textbf{EmbedKGQA}~\cite{saxena2020improving} takes KGQA as a link prediction task and incorporates knowledge graph embeddings~\cite{bordes2013translating,trouillon2016complex} to help predict the answer.

\subsection{Implementations}\label{sec:implementation}
We added \textit{reversed relations} into the relation graph, leading to double size of predicates and triples.
For the text form, we exchanged the placeholder <sub> and <obj> as the reversed relation, \textit{e.g.}, \textit{<sub> co-founded the <obj>} is converted to \textit{<obj> co-founded the <sub>}.

For the experiments of MetaQA, we set the step number $T=3$. We used bi-directional GRU~\cite{chung2014gru} as the question encoder, and set the hidden dimension as $1024$. The projecting function $f^t$ was a stack of linear layer and Tanh layer. The involved MLPs were implemented as simple linear layers.
For the text form, we used another bi-directional GRU as the relation encoder.
The threshold $\tau$ was set to $0.7$ and $\omega$ was set to $400$.
Since the question hop is provided in MetaQA, we used the golden hop number as an auxiliary objective to help learn the hop distribution $\mathbf{c}$.
We computed the cross entropy loss and added it into Equation~\ref{eq:objective} after multiplying a factor of $0.01$.
The model was optimized using RAdam~\cite{liu2019radam} with a learning rate $0.001$ for 20 epochs, which took several hours for the label form and about one day for the text form on a single GPU of NVIDIA 1080Ti.

For the experiments of WebQSP and CompWebQ, we set the step number $T=2$.
We used a pretrained BERT~\cite{devlin2018bert} as the question encoder and finetuned its parameters on our task.
There is no hop annotations so we did not use the auxiliary loss.
Other settings are the same as MetaQA.

\section{Results}

\begin{table*}[ht]
\small
\centering
\begin{tabular}{l|ccc|c|c}
\hline
\multirow{2}{*}{\textbf{Model}} & \multicolumn{3}{c|}{\textbf{MetaQA}} & \multirow{2}{*}{\textbf{WebQSP}} & \multirow{2}{*}{\textbf{CompWebQ}} \\ 
\cline{2-4}
   & \textbf{1-hop}   &   \textbf{2-hop}   & \textbf{3-hop}   &    \\
\hline
KVMemNN~\cite{miller2016key}  &  95.8  &  25.1  &  10.1  &   46.7  &  21.1  \\
VRN~\cite{zhang2017variational}   &  \textbf{97.5}   &  89.9   &   62.5  &   -  & - \\
GraftNet~\cite{sun2018open}  &  97.0  &  94.8  &  77.7  &  66.4  &  32.8 \\
PullNet~\cite{sun2019pullnet}  &  97.0  &  99.9  &  91.4  &  68.1  &  47.2  \\
SRN~\cite{qiu2020stepwise}   &   97.0  &   95.1  &   75.2  &  -  &  - \\
ReifKB~\cite{cohen2020scalable}  &  96.2  &  81.1  &  72.3  &  52.7  &  - \\
EmbedKGQA~\cite{saxena2020improving}  &  \textbf{97.5}  &  98.8  &  94.8  &  66.6  &  -  \\
TransferNet (Ours)  &  \textbf{97.5}  &  \textbf{100}  &  \textbf{100}  &  \textbf{71.4}  &  \textbf{48.6}  \\
\hline
\end{tabular}
\caption{\label{tab:results_label} Hits@1 results of the label-formed datasets. TransferNet achieves 100\% accuracy in the 2-hop and 3-hop questions of MetaQA. On WebQSP and CompWebQ it also outperforms baseline models by a large margin.}
\end{table*}

\begin{table*}[ht]
\small
\centering
\begin{tabular}{l|ccc|ccc}
\hline
\multirow{2}{*}{\textbf{Model}} & \multicolumn{3}{c|}{\textbf{MetaQA Text}} & \multicolumn{3}{c}{\textbf{MetaQA Text + 50\% Label}} \\ 
\cline{2-7}
   & \textbf{1-hop}   &   \textbf{2-hop}   & \textbf{3-hop} & \textbf{1-hop}   &   \textbf{2-hop}   & \textbf{3-hop}  \\
\hline
KVMemNN~\cite{miller2016key}  &  75.4  &  7.0  &  19.5  &  75.7  &  48.4  &  35.2  \\
GraftNet~\cite{sun2018open}  &  82.5  &  36.2  &  40.2  &  91.5  &  69.5  &  66.4  \\
PullNet~\cite{sun2019pullnet}  &  84.4  &  81.0  &  78.2  &  92.4  &  90.4  &  85.2 \\
TransferNet (Ours)  &  \textbf{95.5}  &  \textbf{98.1}  &  \textbf{94.3}  &  \textbf{96.0}  &  \textbf{98.5}  &  \textbf{94.7}    \\
\hline
\end{tabular}
\caption{\label{tab:results_text} Hits@1 results on MetaQA of the text form and mixed form. }
\end{table*}

\subsection{Results on Label-Formed Graph}

Table~\ref{tab:results_label} compares different models on label-formed datasets.
TransferNet performs perfectly in the 2-hop and 3-hop questions of MetaQA, that is, achieving 100\% accuracy.
As for the 1-hop questions of MetaQA, TransferNet achieves 97.5\%, on a par with previous models like VRN and EmbedKGQA.
We analyze the wrong cases of 1-hop and find that the errors are caused by the ambiguity of entities.
For example, the question \textit{who acted in The Last of the Mohicans} asks the actors of the movie \textit{The Last of the Mohicans}.
In the knowledge graph there are two movies with this name, one released in 1936 and the other released in 1920.
Our model outputs the actors of both movies, whereas the MetaQA dataset only considers the actors of the 1920 one as golden answer, causing an inevitable mismatch.
Previous work's performance should also suffer from this dataset fault.
In the questions of 2-hop and 3-hop, the ambiguity is mostly eliminated by the relation restrictions.
Therefore, TransferNet can achieve 100\% accuracy.
We can say that the label-formed MetaQA dataset has been nearly solved by our TransferNet.

WebQSP is more challenging than MetaQA, because it has a much more predicates and triples yet much less training examples.
TransferNet achieves 71.4\% accuracy, beating previous state-of-the-art models (68.1\%) by a large margin, implying that it is well qualified for large-scale knowledge base.

On the CompWebQ dataset, we compare the results with \citet{sun2019pullnet} on the dev set. TransferNet achieves 48.6\% accuracy, still better than PullNet (47.2\%).

\begin{figure*}[t]
\centering
\includegraphics[width=0.9\linewidth]{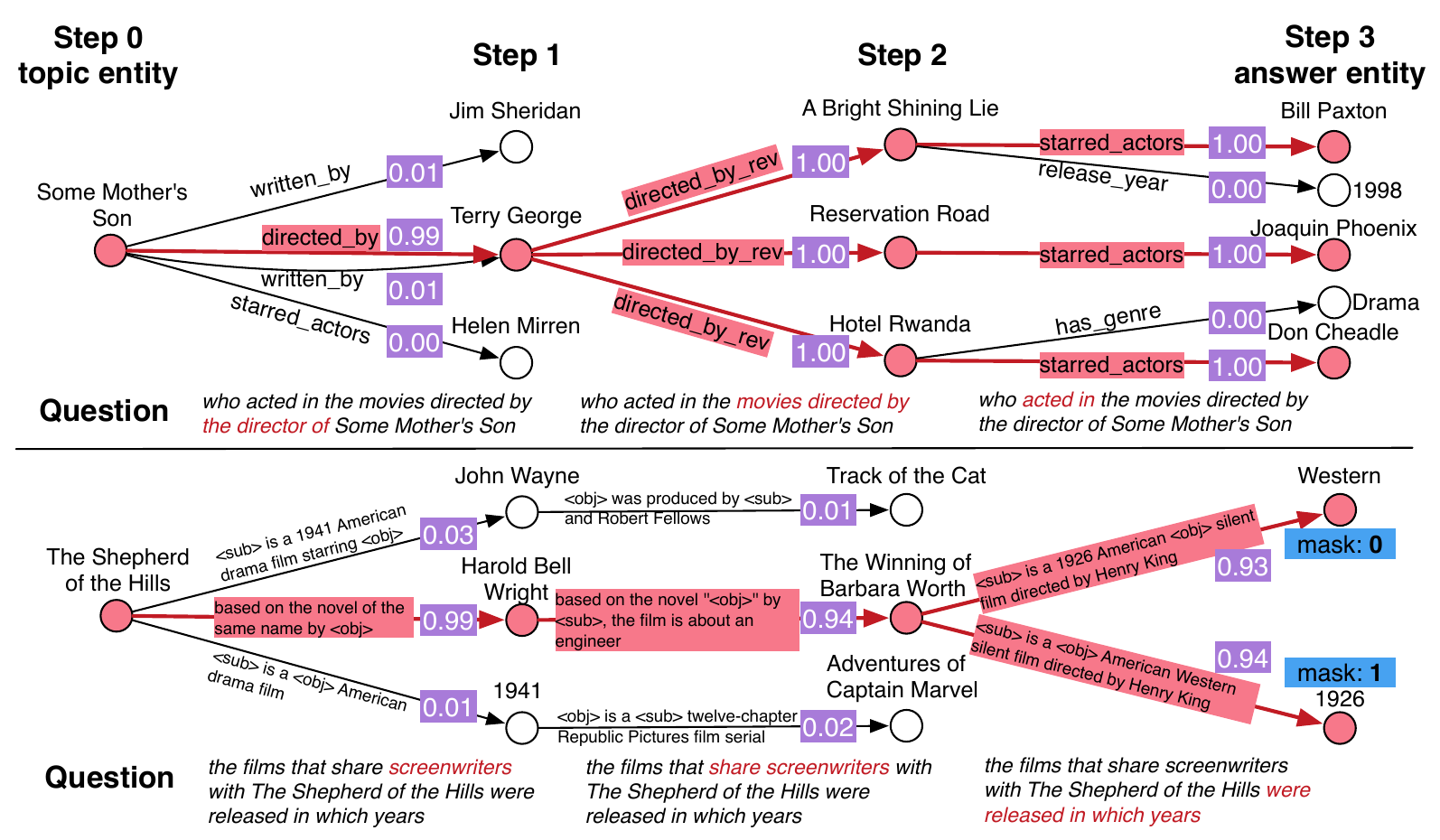}
\caption{Reasoning process of 3-hop questions. The top is in label form, where the suffix ``\_rev'' means reverse relation. The bottom is in text form, where ``mask'' in blue means the language mask. We show the relation scores in purple and highlight the activated entities and relations (score > 0.8) and words (score > 0.05) in red.}
\label{fig:visual}
\end{figure*}

\subsection{Results on Text-Formed Graph}

In Table~\ref{tab:results_label} we compare TransferNet with state-of-the-art models that are able to handle text-formed relations.
We can see that TransferNet significantly outperforms previous models.
Especially for questions of 2-hop and 3-hop, we improve the accuracy from 81.0\% to 98.1\% and from 78.2\% to 94.3\% respectively.
PullNet and GraftNet both infer the answer by aggregating the graph features implicitly, and thus cannot provide the intermediate relation path.
Compared with them, TransferNet not only has a superior performance, but also has a better interpretability (see Sec.\ref{sec:interpretability}).

Besides the pure text form, we also compare the \textit{mixed form} following \cite{sun2018open,sun2019pullnet}.
That is, randomly selecting 50\% of the label-formed triples and add them into the text-formed relation graph.
In this setting, we simply consider the predicates as sentences containing just one word, and use the relation encoder (see Sec.\ref{sec:text_form}) to process them.
These 50\% labels slightly improve the performance of TransferNet over the pure text form (about 0.4\%), because some relations are missing in the text corpus.
Compared with PullNet, TransferNet is still in the lead by a large gap (85.2\% v.s. 94.7\%).

\begin{table}[ht]
\small
\centering
\begin{tabular}{l|c|c}
\hline
 &  Label Form  &  Text Form \\
\hline
TransferNet  &  99.4  &  95.8  \\
w/o score truncation   &  94.7  &  75.3 \\
w/o language mask   &  -  &  62.1  \\
w/o auxiliary loss  &  98.6  &  94.7 \\
\hline
\end{tabular}
\caption{\label{tab:ablation} Ablation study on MetaQA. We show the average hits@1 of different hops.}
\end{table}

\subsection{Ablation Study}

Table~\ref{tab:ablation} shows results of ablation study. 
We can see that the score truncation and language mask are both important, especially for the text form.
As stated in Sec.~\ref{sec:language_mask}, the language mask is not needed in the label form.
The auxiliary loss (see Sec.~\ref{sec:implementation}) slightly improves the performance because it helps the learning of hop attention.

\subsection{Interpretability}\label{sec:interpretability}
We visualize the intermediate results of TransferNet for two 3-hop questions in Figure~\ref{fig:visual}.
The entities and relations whose score is larger than $0.8$ are highlighted in red.
The top question is aimed at the label-formed relation graph.
The activated predicates for three hops are \textit{directed\_by}, \textit{directed\_by\_rev}, and \textit{starred\_actors} respectively, where the suffix \textit{\_rev} means reverse relation.
The bottom question is aimed at the text form.
At step 1, TransferNet tries to find the \textit{screenwriter} of the topic movie, and activates the relation whose textual description is ``\textit{based on the novel of the same name by <obj>}''.
At step 2, the movie written by \textit{Harold Bell Wright} is found.
At step 3, we aim to find the movie's release year.
But since the text descriptions of \textit{Western} (which is the movie's genre) and \textit{1926} are very similar, both of these two entities are activated.
Here the proposed language mask successfully filters the wrong answers out.

\subsection{Model Efficiency}
\begin{figure}[ht]
\centering
\includegraphics[width=\linewidth]{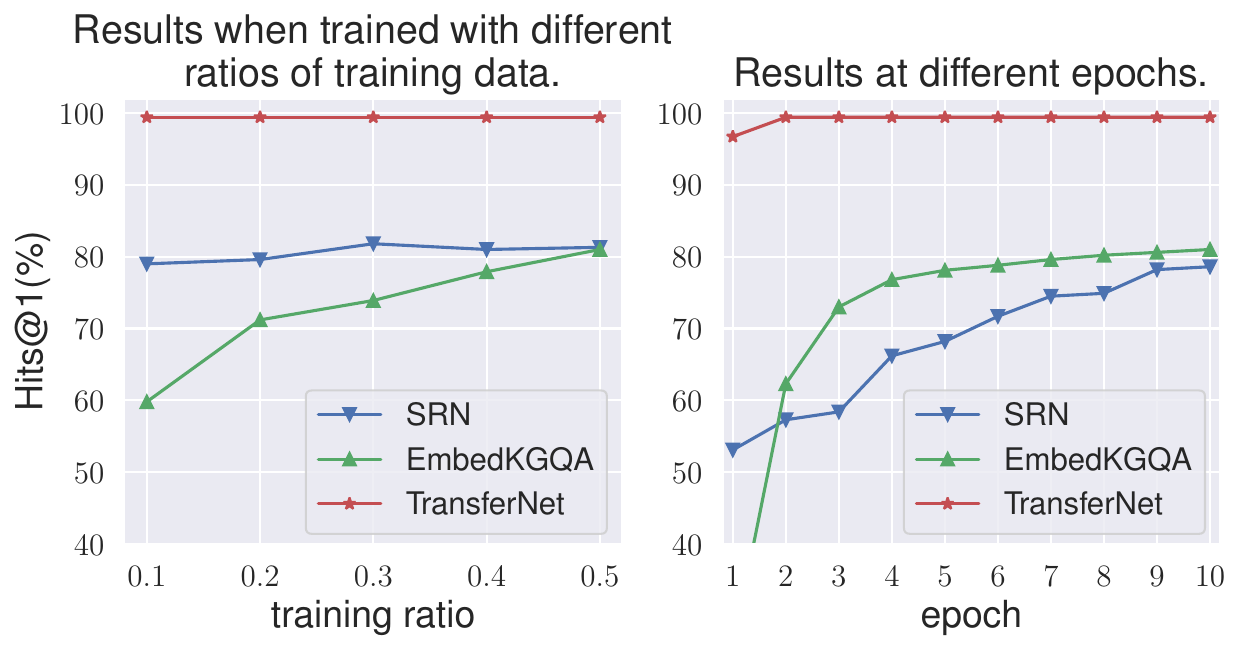}
\caption{Comparison of data efficiency (left) and convergency speed (right) on label-formed MetaQA.}
\label{fig:two_results}
\end{figure}

Figure~\ref{fig:two_results} shows the average hits@1 on the label form of MetaQA when the models are trained with partial training examples (left) and at different epochs (right).
We can see that TransferNet is very data-efficient and converges very fast.
With only 10\% training data, it still achieves the same performance as the entire training set.
And it only needs two epochs to reach the optimal results.

\section{Conclusions}
We proposed TransferNet, an effective and transparent framework for multi-hop QA over knowledge graph or text-formed relation graph.
It achieved 100\% accuracy on 2-hop and 3-hop questions of label-formed MetaQA, nearly solving the dataset.
On the more challenging WebQSP, CompWebQ and text-formed MetaQA, it also outperforms other state-of-the-art models significantly.
Qualitative analysis shows the good interpretability of TransferNet.

\section*{Acknowledgments}
This work is supported by the NSFC Key Project (U1736204), grants from the Institute for Guo Qiang, Tsinghua University (2019GQB0003), Beijing Academy of Artificial Intelligence, Huawei Inc, and MOE AcRF Tier 2.

\bibliography{reference}

\begin{thebibliography}{39}
\expandafter\ifx\csname natexlab\endcsname\relax\def\natexlab#1{#1}\fi

\bibitem[{Berant et~al.(2013)Berant, Chou, Frostig, and
  Liang}]{berant2013semantic}
Jonathan Berant, Andrew Chou, Roy Frostig, and Percy Liang. 2013.
\newblock Semantic parsing on freebase from question-answer pairs.
\newblock In \emph{EMNLP}.

\bibitem[{Bollacker et~al.(2008)Bollacker, Evans, Paritosh, Sturge, and
  Taylor}]{Freebase}
Kurt Bollacker, Colin Evans, Praveen Paritosh, Tim Sturge, and Jamie Taylor.
  2008.
\newblock Freebase: a collaboratively created graph database for structuring
  human knowledge.
\newblock In \emph{SIGMOD}.

\bibitem[{Bordes et~al.(2015)Bordes, Usunier, Chopra, and
  Weston}]{bordes2015large}
Antoine Bordes, Nicolas Usunier, Sumit Chopra, and Jason Weston. 2015.
\newblock Large-scale simple question answering with memory networks.
\newblock \emph{arXiv preprint arXiv:1506.02075}.

\bibitem[{Bordes et~al.(2013)Bordes, Usunier, Garcia-Duran, Weston, and
  Yakhnenko}]{bordes2013translating}
Antoine Bordes, Nicolas Usunier, Alberto Garcia-Duran, Jason Weston, and Oksana
  Yakhnenko. 2013.
\newblock Translating embeddings for modeling multi-relational data.
\newblock In \emph{Advances in neural information processing systems}, pages
  2787--2795.

\bibitem[{Chen et~al.(2017)Chen, Fisch, Weston, and Bordes}]{chen2017reading}
Danqi Chen, Adam Fisch, Jason Weston, and Antoine Bordes. 2017.
\newblock Reading wikipedia to answer open-domain questions.
\newblock In \emph{ACL}, pages 1870--1879.

\bibitem[{Chung et~al.(2014)Chung, Gulcehre, Cho, and Bengio}]{chung2014gru}
Junyoung Chung, Caglar Gulcehre, KyungHyun Cho, and Yoshua Bengio. 2014.
\newblock Empirical evaluation of gated recurrent neural networks on sequence
  modeling.
\newblock \emph{arXiv preprint arXiv:1412.3555}.

\bibitem[{Cohen et~al.(2020)Cohen, Sun, Hofer, and Siegler}]{cohen2020scalable}
William~W Cohen, Haitian Sun, R~Alex Hofer, and Matthew Siegler. 2020.
\newblock Scalable neural methods for reasoning with a symbolic knowledge base.
\newblock \emph{arXiv preprint arXiv:2002.06115}.

\bibitem[{Devlin et~al.(2018)Devlin, Chang, Lee, and
  Toutanova}]{devlin2018bert}
Jacob Devlin, Ming-Wei Chang, Kenton Lee, and Kristina Toutanova. 2018.
\newblock Bert: Pre-training of deep bidirectional transformers for language
  understanding.
\newblock \emph{arXiv preprint arXiv:1810.04805}.

\bibitem[{Ding et~al.(2019)Ding, Zhou, Chen, Yang, and
  Tang}]{ding2019cognitive}
Ming Ding, Chang Zhou, Qibin Chen, Hongxia Yang, and Jie Tang. 2019.
\newblock Cognitive graph for multi-hop reading comprehension at scale.
\newblock In \emph{ACL}, pages 2694--2703.

\bibitem[{Dua et~al.(2019)Dua, Wang, Dasigi, Stanovsky, Singh, and
  Gardner}]{dua2019drop}
Dheeru Dua, Yizhong Wang, Pradeep Dasigi, Gabriel Stanovsky, Sameer Singh, and
  Matt Gardner. 2019.
\newblock Drop: A reading comprehension benchmark requiring discrete reasoning
  over paragraphs.
\newblock In \emph{NAACL-HLT}, pages 2368--2378.

\bibitem[{Fang et~al.(2019)Fang, Sun, Gan, Pillai, Wang, and
  Liu}]{fang2019hierarchical}
Yuwei Fang, Siqi Sun, Zhe Gan, Rohit Pillai, Shuohang Wang, and Jingjing Liu.
  2019.
\newblock Hierarchical graph network for multi-hop question answering.
\newblock \emph{arXiv preprint arXiv:1911.03631}.

\bibitem[{Guo et~al.(2018)Guo, Tang, Duan, Zhou, and Yin}]{guo2018dialog}
Daya Guo, Duyu Tang, Nan Duan, Ming Zhou, and Jian Yin. 2018.
\newblock Dialog-to-action: Conversational question answering over a
  large-scale knowledge base.
\newblock In \emph{Advances in Neural Information Processing Systems}.

\bibitem[{Jiang et~al.(2019)Jiang, Wu, and Jiang}]{jiang2019freebaseqa}
Kelvin Jiang, Dekun Wu, and Hui Jiang. 2019.
\newblock Freebaseqa: a new factoid qa data set matching trivia-style
  question-answer pairs with freebase.
\newblock In \emph{NAACL-HLT}, pages 318--323.

\bibitem[{Joshi et~al.(2017)Joshi, Choi, Weld, and
  Zettlemoyer}]{joshi2017triviaqa}
Mandar Joshi, Eunsol Choi, Daniel~S Weld, and Luke Zettlemoyer. 2017.
\newblock Triviaqa: A large scale distantly supervised challenge dataset for
  reading comprehension.
\newblock In \emph{ACL}, pages 1601--1611.

\bibitem[{Kipf and Welling(2016)}]{kipf2016semi}
Thomas~N Kipf and Max Welling. 2016.
\newblock Semi-supervised classification with graph convolutional networks.
\newblock \emph{arXiv preprint arXiv:1609.02907}.

\bibitem[{Lan et~al.(2019)Lan, Chen, Goodman, Gimpel, Sharma, and
  Soricut}]{lan2019albert}
Zhenzhong Lan, Mingda Chen, Sebastian Goodman, Kevin Gimpel, Piyush Sharma, and
  Radu Soricut. 2019.
\newblock Albert: A lite bert for self-supervised learning of language
  representations.
\newblock \emph{arXiv preprint arXiv:1909.11942}.

\bibitem[{Liang et~al.(2017)Liang, Berant, Le, Forbus, and
  Lao}]{liang2017neural}
Chen Liang, Jonathan Berant, Quoc Le, Kenneth Forbus, and Ni~Lao. 2017.
\newblock Neural symbolic machines: Learning semantic parsers on freebase with
  weak supervision.
\newblock In \emph{ACL}.

\bibitem[{Liu et~al.(2020)Liu, Jiang, He, Chen, Liu, Gao, and
  Han}]{liu2019radam}
Liyuan Liu, Haoming Jiang, Pengcheng He, Weizhu Chen, Xiaodong Liu, Jianfeng
  Gao, and Jiawei Han. 2020.
\newblock On the variance of the adaptive learning rate and beyond.
\newblock In \emph{ICLR}.

\bibitem[{Miller et~al.(2016)Miller, Fisch, Dodge, Karimi, Bordes, and
  Weston}]{miller2016key}
Alexander Miller, Adam Fisch, Jesse Dodge, Amir-Hossein Karimi, Antoine Bordes,
  and Jason Weston. 2016.
\newblock Key-value memory networks for directly reading documents.
\newblock In \emph{EMNLP}, pages 1400--1409.

\bibitem[{Petrochuk and Zettlemoyer(2018)}]{petrochuk2018simplequestions}
Michael Petrochuk and Luke Zettlemoyer. 2018.
\newblock Simplequestions nearly solved: A new upperbound and baseline
  approach.
\newblock In \emph{EMNLP}, pages 554--558.

\bibitem[{Qiu et~al.(2020)Qiu, Wang, Jin, and Zhang}]{qiu2020stepwise}
Yunqi Qiu, Yuanzhuo Wang, Xiaolong Jin, and Kun Zhang. 2020.
\newblock Stepwise reasoning for multi-relation question answering over
  knowledge graph with weak supervision.
\newblock In \emph{WSDM}, pages 474--482.

\bibitem[{Rajpurkar et~al.(2016)Rajpurkar, Zhang, Lopyrev, and
  Liang}]{rajpurkar2016squad}
Pranav Rajpurkar, Jian Zhang, Konstantin Lopyrev, and Percy Liang. 2016.
\newblock Squad: 100,000+ questions for machine comprehension of text.
\newblock In \emph{EMNLP}, pages 2383--2392.

\bibitem[{Saha et~al.(2019)Saha, Ansari, Laddha, Sankaranarayanan, and
  Chakrabarti}]{saha2019complex}
Amrita Saha, Ghulam~Ahmed Ansari, Abhishek Laddha, Karthik Sankaranarayanan,
  and Soumen Chakrabarti. 2019.
\newblock Complex program induction for querying knowledge bases in the absence
  of gold programs.
\newblock \emph{Transactions of the Association for Computational Linguistics}.

\bibitem[{Saxena et~al.(2020)Saxena, Tripathi, and
  Talukdar}]{saxena2020improving}
Apoorv Saxena, Aditay Tripathi, and Partha Talukdar. 2020.
\newblock Improving multi-hop question answering over knowledge graphs using
  knowledge base embeddings.
\newblock In \emph{ACL}, pages 4498--4507.

\bibitem[{Sun et~al.(2019)Sun, Bedrax-Weiss, and Cohen}]{sun2019pullnet}
Haitian Sun, Tania Bedrax-Weiss, and William Cohen. 2019.
\newblock Pullnet: Open domain question answering with iterative retrieval on
  knowledge bases and text.
\newblock In \emph{EMNLP-IJCNLP}, pages 2380--2390.

\bibitem[{Sun et~al.(2018)Sun, Dhingra, Zaheer, Mazaitis, Salakhutdinov, and
  Cohen}]{sun2018open}
Haitian Sun, Bhuwan Dhingra, Manzil Zaheer, Kathryn Mazaitis, Ruslan
  Salakhutdinov, and William Cohen. 2018.
\newblock Open domain question answering using early fusion of knowledge bases
  and text.
\newblock In \emph{EMNLP}, pages 4231--4242.

\bibitem[{Talmor and Berant(2018)}]{talmor2018web}
Alon Talmor and Jonathan Berant. 2018.
\newblock The web as a knowledge-base for answering complex questions.
\newblock In \emph{NAACL-HLT}, pages 641--651.

\bibitem[{Trouillon et~al.(2016)Trouillon, Welbl, Riedel, Gaussier, and
  Bouchard}]{trouillon2016complex}
Th{\'e}o Trouillon, Johannes Welbl, Sebastian Riedel, {\'E}ric Gaussier, and
  Guillaume Bouchard. 2016.
\newblock Complex embeddings for simple link prediction.
\newblock ICML.

\bibitem[{Tu et~al.(2020)Tu, Huang, Wang, Huang, He, and Zhou}]{tu2020select}
Ming Tu, Kevin Huang, Guangtao Wang, Jing Huang, Xiaodong He, and Bowen Zhou.
  2020.
\newblock Select, answer and explain: Interpretable multi-hop reading
  comprehension over multiple documents.
\newblock In \emph{AAAI}, pages 9073--9080.

\bibitem[{Vrande{\v{c}}i{\'c} and Kr{\"o}tzsch(2014)}]{Wikidata}
Denny Vrande{\v{c}}i{\'c} and Markus Kr{\"o}tzsch. 2014.
\newblock Wikidata: a free collaborative knowledge base.

\bibitem[{Xu et~al.(2019)Xu, Lai, Feng, and Wang}]{xu2019enhancing}
Kun Xu, Yuxuan Lai, Yansong Feng, and Zhiguo Wang. 2019.
\newblock Enhancing key-value memory neural networks for knowledge based
  question answering.
\newblock In \emph{NAACL-HLT}, pages 2937--2947.

\bibitem[{Yang et~al.(2018)Yang, Qi, Zhang, Bengio, Cohen, Salakhutdinov, and
  Manning}]{yang2018hotpotqa}
Zhilin Yang, Peng Qi, Saizheng Zhang, Yoshua Bengio, William Cohen, Ruslan
  Salakhutdinov, and Christopher~D Manning. 2018.
\newblock Hotpotqa: A dataset for diverse, explainable multi-hop question
  answering.
\newblock In \emph{EMNLP}, pages 2369--2380.

\bibitem[{Yih et~al.(2015)Yih, Chang, He, and Gao}]{yih2015semantic}
Wen-tau Yih, Ming-Wei Chang, Xiaodong He, and Jianfeng Gao. 2015.
\newblock Semantic parsing via staged query graph generation: Question
  answering with knowledge base.
\newblock In \emph{ACL-IJCNLP}, pages 1321--1331.

\bibitem[{Yih et~al.(2016)Yih, Richardson, Meek, Chang, and Suh}]{yih2016value}
Wen-tau Yih, Matthew Richardson, Christopher Meek, Ming-Wei Chang, and Jina
  Suh. 2016.
\newblock The value of semantic parse labeling for knowledge base question
  answering.
\newblock In \emph{ACL}, pages 201--206.

\bibitem[{Zhang et~al.(2017)Zhang, Dai, Kozareva, Smola, and
  Song}]{zhang2017variational}
Yuyu Zhang, Hanjun Dai, Zornitsa Kozareva, Alexander~J Smola, and Le~Song.
  2017.
\newblock Variational reasoning for question answering with knowledge graph.
\newblock \emph{arXiv preprint arXiv:1709.04071}.

\bibitem[{Zhang et~al.(2020)Zhang, Yang, and Zhao}]{zhang2020retrospective}
Zhuosheng Zhang, Junjie Yang, and Hai Zhao. 2020.
\newblock Retrospective reader for machine reading comprehension.
\newblock \emph{arXiv preprint arXiv:2001.09694}.

\bibitem[{Zhao et~al.(2019{\natexlab{a}})Zhao, Xiong, Rosset, Song, Bennett,
  and Tiwary}]{zhao2019transformer}
Chen Zhao, Chenyan Xiong, Corby Rosset, Xia Song, Paul Bennett, and Saurabh
  Tiwary. 2019{\natexlab{a}}.
\newblock Transformer-xh: Multi-evidence reasoning with extra hop attention.
\newblock In \emph{International Conference on Learning Representations}.

\bibitem[{Zhao et~al.(2019{\natexlab{b}})Zhao, Chung, Goyal, and
  Metallinou}]{zhao2019simple}
Wenbo Zhao, Tagyoung Chung, Anuj Goyal, and Angeliki Metallinou.
  2019{\natexlab{b}}.
\newblock Simple question answering with subgraph ranking and joint-scoring.
\newblock In \emph{NAACL-HLT}, pages 324--334.

\bibitem[{Zhou et~al.(2018)Zhou, Huang, and Zhu}]{zhou2018interpretable}
Mantong Zhou, Minlie Huang, and Xiaoyan Zhu. 2018.
\newblock An interpretable reasoning network for multi-relation question
  answering.
\newblock In \emph{COLING}.

\end{thebibliography}
\bibliographystyle{acl_natbib}

\end{document}